\documentclass[letterpaper]{article} % DO NOT CHANGE THIS
\usepackage{aaai23}  % DO NOT CHANGE THIS
\usepackage{times}  % DO NOT CHANGE THIS
\usepackage{helvet}  % DO NOT CHANGE THIS
\usepackage{courier}  % DO NOT CHANGE THIS
\usepackage[hyphens]{url}  % DO NOT CHANGE THIS
\usepackage{graphicx} % DO NOT CHANGE THIS
\def\UrlFont{\rm}  % DO NOT CHANGE THIS
\usepackage{natbib}  % DO NOT CHANGE THIS AND DO NOT ADD ANY OPTIONS TO IT
\usepackage{caption} % DO NOT CHANGE THIS AND DO NOT ADD ANY OPTIONS TO IT
\usepackage{algorithm}
\usepackage{algorithmic}
\usepackage{wrapfig}
\usepackage{color}
\usepackage{makecell}
\usepackage{multirow}
\usepackage{mathrsfs}
\usepackage{amssymb}
\usepackage{amsmath}

\newcommand{\squishlist}[1][$\bullet$]
{
    \begin{list}{#1}
    {
        \setlength{\itemsep}{0pt}
        \setlength{\parsep}{2pt}
        \setlength{\topsep}{2pt}
        \setlength{\partopsep}{0pt}
        \setlength{\leftmargin}{1.5em}
        \setlength{\labelwidth}{1.5em}
        \setlength{\labelsep}{0.5em}
    }
}

\newcommand{\squishend}{\end{list}}

\usepackage{newfloat}
\usepackage{listings}
\DeclareCaptionStyle{ruled}{labelfont=normalfont,labelsep=colon,strut=off} % DO NOT CHANGE THIS
\floatstyle{ruled}
\newfloat{listing}{tb}{lst}{}
\floatname{listing}{Listing}
\title{Future Aware Pricing and Matching for Sustainable On-Demand Ride Pooling}
\author {
    Xianjie Zhang\textsuperscript{\rm 1,\rm 2},
    Pradeep Varakantham\textsuperscript{\rm 2},
    Hao Jiang\textsuperscript{\rm 2}
}
\affiliations {
    \textsuperscript{\rm 1}Key Laboratory for Ubiquitous Network and Service Software of Liaoning Province, School of Software, 
    Dalian University of Technology, China\\
    \textsuperscript{\rm 2}%School of Computing and Information Systems, 
    Singapore Management University, Singapore\\
    xj$\_$zh@foxmail.com, pradeepv@smu.edu.sg, haojiang.2021@phdcs.smu.edu.sg
}

\begin{document}

\maketitle

\begin{abstract}
The popularity of on-demand ride pooling is owing to the benefits offered to customers (lower prices), taxi drivers (higher revenue), environment (lower carbon footprint due to fewer vehicles) and aggregation companies like Uber (higher revenue). To achieve these benefits, two key interlinked challenges have to be solved effectively: (a) pricing -- setting prices to customer requests for taxis; and (b) matching -- assignment of customers (that accepted the prices) to taxis/cars. Traditionally, both these challenges have been studied individually and using myopic approaches (considering only current requests), without considering the impact of current matching on addressing future requests. In this paper, we develop a novel framework that handles the pricing and matching problems together, while also considering the future impact of the pricing and matching decisions. 
In our experimental results on a real-world taxi dataset, we demonstrate that our framework can significantly improve revenue (up to 17\% and on average 6.4\%) in a sustainable manner by reducing the number of vehicles  (up to 14\% and on average 10.6\%) required to obtain a given fixed revenue and the overall distance travelled by vehicles (up to 11.1\% and on average 3.7\%). That is to say, we are able to provide an ideal win-win scenario for all stakeholders (customers, drivers, aggregator, environment) involved by obtaining higher revenue for customers, drivers, aggregator (ride pooling company) while being good for the environment (due to fewer number of vehicles on the road and lesser fuel consumed).
\end{abstract}

\section*{Introduction}
The emergence of Uber, LYFT, DiDi and other taxi aggregation companies have improved the utilization rate of taxis and reduced customers' waiting time. One of the most popular services offered by taxi aggregation companies is the on-demand ride pooling service.  Apart from being one of the most economical transport methods that allows multiple passengers to share a taxi, on-demand ride pooling also provides higher revenue for drivers and taxi aggregation companies, while having fewer taxis on the roads (lower carbon footprint). Effectiveness and efficiency of ride pooling service is determined by the methods used for pricing of customer requests (referred to as the pricing problem) and matching of taxis to groups of customer requests (referred to as the matching problem).
 
There are multiple difficult challenges in solving the matching and pricing problems for ride-pooling systems. First, finding taxis to serve requests (matching) determines the price to be set for requests (pricing) and price set for a request (pricing) determines which requests still remain (as customers can drop the request if the price is too high) for the taxis to serve. In summary, there is a cyclic dependency between pricing and matching, where pricing impacts matching and matching impacts pricing. Second, both pricing and matching decisions can have an impact on future requests and recent work~\cite{shah2020neural} has shown that developing future aware approaches can have a significant impact. Finally, unlike in on-demand ride sharing, on-demand ride pooling requires matching vehicles to trips (combinations of requests), which is no longer a bi-partite matching problem. It is matching on a tripartite graph~\cite{Alonso-Mora462} between requests, trips (combinations of requests) and vehicles, which is a significantly harder problem. 
 
Due to the cyclic dependency between matching and pricing, recent work has acknowledged the importance of simultaneously optimizing pricing and matching~\cite{shah2022icaps,jointpandm} in improving the overall effectiveness and efficiency of on-demand ride sharing systems.  However, most existing works have focused on optimizing either pricing or matching individually \cite{3_p,bimpikis2019spatial,14_p,shah2020neural,shah2022icaps} and usually in a myopic manner (without considering the impact on serving future requests), but recently there has been emphasis on future aware methods~\cite{shah2020neural,Lowalekar2019ZACAZ} that have shown to provide significant benefits. There have been few recent studies on joint optimal pricing and matching decisions, however, they are either not scalable \cite{chen2019dispatching} or rely on unrealistic assumptions about knowing all (current and future) customer requests in advance \cite{chen2019inbede}. Moreover, the above works~\cite{chen2019dispatching,chen2019inbede} are not for the case of ride pooling.

\textbf{To that end, in this paper, we make multiple contributions to simultaneously optimize matching and pricing decisions, while considering the future impact of those decisions for sustainable city scale on-demand ride pooling systems.} Specifically, we make the following contributions:
\begin{itemize}
    \item We provide a precise and formal definition of the overall matching and pricing problem in the context of ride pooling systems. 
    \item We provide a two-layered reinforcement learning approach, that  combines mean field Q-learning for pricing and neural approximate dynamic programming for matching in a synergistic manner to optimize both pricing and matching. We employ different RL approaches for matching and pricing due to the special structure present in the two problems. 
    \item Social impact is typically feasible if there is a win-win situation for all stakeholders involved. In our case, the different stakeholders are customers, drivers, aggregator (Uber etc.) and environment. While customers, drivers and aggregator are more focussed on the revenue and availability of taxis, environment will win if fewer vehicles are on the road and amount of fuel consumed is lower (distance travelled is lower). We provide a detailed experimental section to evaluate our approach on a benchmark simulator that is based on a city scale real-world taxi dataset. Our approach is able to improve up to 17\% on revenue across a wide range of parameter values and reduce the number of vehicles by up to 14\%. As a benchmark, typically 0.5-1\% improvements are considered significant in mobility settings~\cite{Xu2018}.
\end{itemize}

\section*{Pricing and Matching Problem}
\noindent In this problem, we have a set of vehicles $V$, each of which can potentially serve multiple customers (maximum of $c$) simultaneously (e.g., UberPool or GrabShare).  The operator receives a set of requests for rides from customers. Typically, these requests are batched together over a fixed interval (e.g., 60 seconds) and we will refer to this set as $R$. 
\begin{table}
{\small \begin{tabular}{|c|l|}
    \hline
     Symbol & Definition  \\
     \hline
     \hline
     $V$ & Available vehicles in the current time period\\
     \hline
     $R$ & Batched requests in the current time period.\\
     \hline
     $r$ & Variable to denote a request \\
     \hline
     $v$ & Variable to denote a vehicle\\
     \hline
     $f$ & Variable to denote a combination of requests.  \\
     & $R^f$ corresponds to requests in $f$\\
     \hline
     $x^f_v$ & Binary decision variable that is set to 1 if\\ 
     & $v$ is assigned to $f$\\
     \hline
     $\mu$ & Pricing vector for all requests at a time step \\
     \hline
     $o_v^{f}(\mu^f)$ & Expected reward obtained by assigning request \\
     &  combination ${f}$ to $v$ given price vector $\mu^f$\\
     \hline
     $p(.)$ & Price-sensitivity function maps the set of \\
     & prices $\mu$ to the probability of acceptance $p(\mu)$.\\ 
     & It is assumed to be known apriori. For a given \\ 
     &request $r \in R$,  $p_r(\mu_r)$ is the probability of\\ 
     & acceptance of the customer at the price $\mu_r$.\\
     \hline
     $C(.)$ & Matching constraints required for \\
     & feasible matching \\
     \hline
     $a_v$ & Multiplier on base price for serving a request \\
     \hline
    $\tau$ &  Pick up delay constraint \\
    \hline
    $\lambda$ & Detour delay constraint \\
    \hline
\end{tabular}}
\caption{Symbols and their definition}
\end{table}
The aggregation company then searches the set of available vehicles, $V$ that satisfy quality constraints with respect to wait time and time to destination for the customer requests. Once at least one available vehicle is identified for a request, a price for the customer request is set by the aggregation company, which the customers can decide to accept or reject. There is a trade-off between pricing too high and losing customers vs pricing too low and losing revenue. Our goal is to design a pricing strategy that can ensure that the induced matching maximizes the expected revenue.  We now formally define the `pricing' and `matching' problems. 

\subsection*{Pricing Problem}
The goal of the pricing problem is to compute a pricing vector, $\mu$, which contains a price $\mu_r$ for each customer request, $r \in R$. Formally, the problem of pricing customer requests is formulated as two-level optimization problem:
% \begin{equation}
%     \mu^* = \arg\max\limits_{\mu} \Big[\underset{R_\mu \sim p(\mu)}{\mathbb{E}} \big[\max_{x} (x_v^f \cdot o_v^f) \big] \Big]
% \end{equation}
\begin{equation}
     \mu^* = \arg\max\limits_{\mu} \Big[ \max_{X \in C(R)} \Big(\underset{x_v^f \in X}{\sum} x_v^f \cdot o_v^{{f}}(\mu^f)\big] \Big) \Big]
     \label{eq:1}
\end{equation}
While the outer maximization is for computing the best pricing vector for all requests, $R$, the inner maximization is for finding the matching strategy that will maximize the expected revenue (immediate or long term) given the pricing strategy.  $o_v^{{f}}(\mu^f)$ is the expected immediate  (or long term) reward given the pricing vector, $\mu^f$ for requests in the request combination, $f$.  There are multiple key challenges in solving this two-level optimization while being future aware (i.e., considering potential future requests):
\squishlist
\item There are exponentially many pricing strategies and computing the outer maximum is expensive. 
\item From Equation~\ref{eq:1}, pricing and matching are dependent on each other. Pricing affects the active requests (i.e., ones for which customers accept the price) and hence affects the matching process. The matching determines which vehicle is assigned to a request and correspondingly the price associated with the request. 
\item The matching and pricing at a time step have an impact on the available vehicles (and potentially requests) at the next time step. 
\squishend

\subsection*{Matching Problem}

The goal of matching is to create an assignment of vehicles to requests such that the objective $o$ is maximized subject to constraints on matchings. This problem can be seen as a `maximum weight bipartite matching' in the un-shared case but, bipartite matching doesn't capture the structure of the matching problem when multiple requests can be matched to a single vehicle. For example, consider  two requests A and B that can be served by Vehicle 1 individually. However, the vehicle cannot serve them together as it would lead to an unacceptable amount of detours; these kinds of combinatorial constraints cannot be captured in a bipartite graph.
The way to solve this problem is to create an intermediate representation called a `trip' which is a combination of requests \cite{Alonso-Mora462} represented using $f$. Then, requests are mapped to vehicles via trips such that both vehicles and requests can be assigned at most one trip. Each trip and vehicle mapping has an associated weight of $o^f_v$, and the solution to the matching problem can be seen as a maximum weight matching in this resulting tripartite graph. We write this optimization using the equation:
\begin{equation}
    X^* = \arg\max_{X \in C(R)} \sum_{x^f_v \in X} x^f_v \cdot o^f_v(\mu^f) \label{eqn:2}
\end{equation}
The matching operation is performed by solving an Integer Linear Program. The specific formulation is given below. While there is an exponentially large number of trips in the worst case, \cite{Alonso-Mora462} provided an approximation method for generating feasible trips that work well in practice. This set of feasible trips, ${\mathcal F}$ contains requests that satisfies quality constraints (described in the next section). 
\begin{align*}
\max \quad & \sum_{x^f_v} x^f_v * o^f_v(\mu^f) \\
s.t. \quad & \sum_{v \in V} x^f_v = 1 ::: \forall f ;; \sum_{f | r \in f} \sum_{v \in V^r} x^f_v \leq  1 ::: \forall r \\
& x^f_v \in \{0,1\} ::: \forall t, v
\end{align*}
The constraints in the above Integer Linear Program (ILP) are referred to as $C (.)$ in Equation~\ref{eqn:2}.

\noindent \textbf{Matching Objective}, $\boldsymbol{o_v^f(\mu^f)}$\textbf{:} 
In our formulation, we abstract away the objective in terms of $o^f_v(\mu^f)$, but concretely it represents the system-level objective that the operator wants to maximise. This could be an obvious goal like profit or revenue, but may also be a way to incorporate future information \cite{shah2020neural} or even fairness metrics \cite{lesmana2019balancing}. All these objectives can be modeled as a linear function of the revenue of the trip.\\
\noindent \emph{Profit:} $$o^f_v (\mu^f) = [\, \sum_{r \in R^f} p^r(\mu^r)\cdot \mu^r ] - {cost}^f_v$$ where ${cost}^f_v$ is a constant denotes the marginal increase in cost incurred by serving a trip $f$ with vehicle $v$.\\
\noindent \emph{NeurADP \cite{shah2020neural}:} $$o^f_v (\mu^f) = [\, \sum_{r \in  R^f} p^r(\mu^r) \cdot \mu^r] + \gamma \cdot \mathbb{V}^f_v$$ where the second term $\gamma \cdot \mathbb{V}^f_v$ is a learned constant. While we look at determining the optimal pricing and matching for a given batching interval in this paper, NeurADP attempts to match riders to drivers such that it is optimal across multiple batching intervals. Their solution takes the form of adding a `future value' $\gamma \cdot \mathbb{V}^f_v$ to the matching objective in every batching interval and fits neatly into our generalisation.\\
\noindent \emph{Historical Earnings \cite{lesmana2019balancing}:} 
$$o^f_v (\mu^f) = [\, \sum_{r \in  R^f} \mu^r\cdot p^r(\mu^r)] + hist_v$$ where the second term is a constant that denotes the historical earnings for a given driver. In the paper, they attempt to even out driver earnings by adding the driver's historical earnings to the objective.\\

As a result, we focus on maximising a set of general objectives that can be written as a linear function of a trip's revenue ($\alpha^f_v$ and $\beta^f_v$ are constants):
\begin{equation} \label{eqn:linearmatch_}
    o^f_v (\mu^f) = \alpha^f_v \cdot {\sum}_{r \in R^f} p^r(\mu^r) \cdot \mu^r + \beta^f_v
\end{equation}
The overall objective is combinatorial in the requests because each $o^f_v$ component depends on the source and destination of all the requests in trip $f$, as well as the existing trajectory of vehicle $v$.

\section*{Related Work}
{\small \begin{table*}[ht]
\centering
\setlength\tabcolsep{3pt}
\begin{tabular}{|l|l|l|l|l|l|l|}
\hline
 & \textbf{Matching} & \textbf{Pricing} & \textbf{Future Demand} & \textbf{Capacity} & \textbf{Request} \\
  & &  &  \textbf{and Price-Sensitivity} & \textbf{of Vehicles} & \textbf{Processing} \\
 \hline
 Banerjee et al.~\citeyear{banerjee2015pricing} & Not optimised & optimised & Known  Distribution & Unit  Capacity & Region based \\
 Banerjee et al.~\citeyear{banerjee2016dynamic} &  &  &  &   & Pricing\\
 Ma et al.~\citeyear{bimpikis2019spatial}&  &  &  &   & Sequential (Matching) \\
  \hline
  Ma et al.~\citeyear{ma2013t}    & optimised & Not optimised  &  No  Information& Multi-Capacity & Batched (Matching)  \\
Zheng et al.~\citeyear{zheng2018order}&  &  &  &   &   \\
   \hline
    Chen et al.~\citeyear{chen2019inbede}    & optimised & optimised & Known Distribution & Unit Capacity & Batched(Matching)  \\
        &  &  &  &  &  Sequential (Pricing) \\
       \hline
  Ma et al.~\citeyear{ma2019spatio}   & optimised & optimised & Exact Information  & Unit Capacity  & Batched  \\
    \hline
\textbf{Our Work} & \textbf{optimised} & \textbf{optimised} & \textbf{Known Distribution} & \textbf{Multi-Capacity}  & \textbf{Batched}  \\
Shah et al.~\citeyear{shah2022icaps} &  &  &  &   &   \\

\hline
\end{tabular}
\caption{Summary of Differences Between Our Work and Related Work}
\label{table:relwork}
\end{table*}}
\noindent The existing work for optimising ride-sharing systems can be categorized based on different dimensions, as shown in Table \ref{table:relwork}. These dimensions include how the pricing and matching decisions are made, the capacity of the vehicles considered, sequential (one by one) or batched (considering all active requests together) processing of requests and the amount of future information available to the algorithm. 

As seen in the table, most of the existing work for ride-sharing systems has studied the decision-making in isolation, i.e., they either perform  matching by assuming a heuristic or fixed pricing~\cite{ma2013t,zheng2018order}, or focus on pricing decisions and ignore the optimisation for matching~\cite{banerjee2015pricing,bimpikis2019spatial}. Most existing works also consider sequential processing of requests, for at least one of the decision-making components. The sequential solution is faster to compute but is typical of poorer quality than the batched solution as it has less information available to make a decision~\cite{uberblog}.

Banerjee et al. \cite{banerjee2015pricing, banerjee2016dynamic} provide a threshold-based pricing mechanism where the platform raises the price whenever the available vehicles in the region fall below a threshold. Their focus is on studying the theoretical properties of such threshold-based policies. \citet{bimpikis2019spatial} provide a spatial pricing scheme for ride-sharing markets where they set a price for each region. As opposed to our work, these works do not focus on optimising matching decisions, and heuristically match the customer with unit-capacity vehicles available in the region. However, while these region-based simplifications may work well in the unit-capacity case, matching in the multi-capacity case is much more complicated and cannot be approximated in the same way.

\citet{ozkan2020joint} theoretically prove that optimising only along one dimension is not optimal in general and joint pricing and matching optimisation can significantly increase performance. Unlike this paper, however, they do not provide an explicit algorithm, but rather focus on deriving conditions in which optimising only pricing or matching individually can help.

There has been some research on providing algorithms for jointly optimising pricing and matching decisions for these systems. \citet{ma2019spatio} provide a spatio-temporal pricing mechanism that provides competitive equilibrium prices and can achieve a higher revenue than a myopic pricing scheme. They make an unrealistic assumption that all the information about future requests is exactly known. On the other hand, the only assumption made about requests in this paper is that we know the price-sensitivity of the customers making the request.

\citet{chen2019inbede,chen2019dispatching} provide frameworks for jointly optimising pricing and matching decisions. However, these formulations either are not suitable to the multi-capacity setting or are not scalable. As shown in our experimental results, our approach can provide decisions in real-time for multi-capacity ride pooling in a city scale problem.

The key difference in our approach and \cite{shah2022icaps} is that we employ future aware joint matching and pricing, while Shah {\em et al.} only employ myopic pricing. 

\section*{Methodology}\label{joint-matching-and-pricing}

In this section, we describe our contributions in solving the meta optimization problem of Equation~\ref{eq:1} while considering the potential future impact of pricing and matching decisions. A brute-force approach would be to employ a centralized Reinforcement Learning method that makes joint pricing and matching decisions. However, such an approach is not viable and suitable, for multiple reasons. First, there are thousands of vehicles and a few hundred requests at each time step, so the complexity of even solving the single step meta optimization of Equation~\ref{eq:1} is very significant. Second, matching decisions are dependent on pricing decisions, so searching for a joint policy that makes both decisions simultaneously can result in searching over many unnecessary policies.  
Finally, matching problem has a specific structure where transition uncertainty (arising due to customer requests) is external to the action space (matching decisions). Similarly, the pricing problem for a vehicle has a specific structure, where the pricing strategy is primarily dependent on the vehicle neighborhood  (requests and vehicles in the neighborhood). A centralized Reinforcement learning method will not be able to exploit such structure. 
\begin{figure}
    \centering
    \includegraphics[width=3.3in,height=2in]{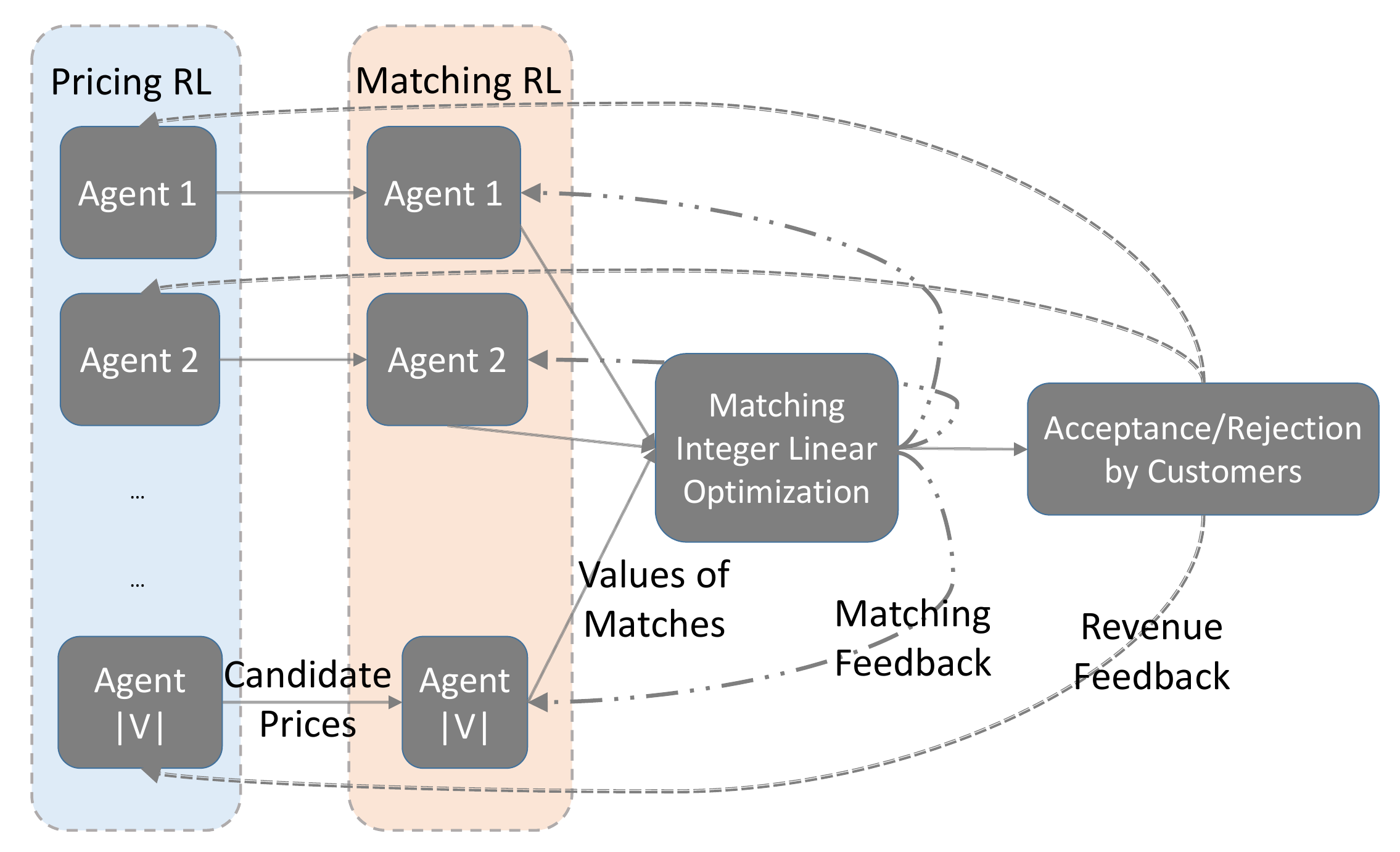}
    \caption{Two layer RL based framework for future aware pricing and matching in on-demand ride pooling}
    \label{fig:2}
\end{figure}
To that end, we provide a two-layered Reinforcement Learning approach for each vehicle, as shown in Figure~\ref{fig:2}. Given the set of requests at a given time step, we have the following overall framework:
\squishlist
\item \textbf{Pricing RL} for each vehicle/agent (referred to as Agent 1 Pricing RL etc. in the figure) is responsible for computing candidate prices for neighborhood requests of that vehicle;
\item Based on the computed candidate prices, \textbf{Matching RL} for each vehicle (referred to as Agent 1 Matching RL, etc.) computes the future value for different possible matches; 
\item Using the future values of different matches from all vehicles, a centralized linear integer optimization program (referred to as the Matching Integer Linear Optimization in the figure) computes the best feasible matching of vehicles to requests; 
\item Based on the assignment output ("Matching Feedback" dotted line in the figure) by the integer optimization program, weights in the deep neural network (in Matching RL) are used for estimating the values of different matches (actions in Matching RL) are updated for the different vehicles; and  
\item Finally, customers can accept or reject the price (if high) and this feedback is used to update the weights in the deep network for estimating the values of different price factors (actions) in Pricing RL for different vehicles. 
\squishend

 %Figure~\ref{fig:Steps} shows the key steps in execution of the overall framework and we now describe our 2 key contributions corresponding to pricing RL ( Figure~\ref{fig:Steps}(b)) and matching and final pricing (Figure~\ref{fig:Steps}(c)).

\subsection*{Pricing RL}
One of the key challenges in solving the pricing optimization problem of Equation~\ref{eq:1} is the maximization over a combinatorial solution space (of pricing vectors). In the real world taxi setup considered in our experiments, the number of requests per minute is around 300, so this would be a 300 dimension search space. The price for a request depends on the supply (available taxis)-demand (customer requests) imbalance\footnote{It will also depend on the price sensitivity of customers and we consider the price sensitivity as part of the matching objective.} around the request location and we exploit this observation in providing a scalable solution.  Instead of centrally deciding on a price vector, as shown in Figure~\ref{fig:2}, we compute candidate prices for requests in the neighborhood of a vehicle using the Pricing RL box for that vehicle. 
Specifically, to capture the dependence of pricing on neighborhood vehicles/agents, we use Mean Field Q-Learning (MFQL) in the pricing RL box.  MFQL generates for each vehicle a multiplier on top of the base price for serving requests that can be served by the vehicle. 

Formally, we initially compute a base price $\mu_{0}^r$ for each request, $r$ based on factors such as source, destination, travel distance, etc. The goal of MFQL is to compute a multiplier, $a_v$ for each vehicle in serving a request in its neighborhood, so as to maximize the overall objective (e.g., revenue). The candidate price, $\tilde{\mu}^r_v$ due to vehicle $v$ for request $r$ will be 
$$\tilde{\mu}^r_v = \mu_0^r \cdot a_v$$
$\tilde{\mu}^r = \{  \tilde{\mu}_v^r \}_{v \in V}$ refers to the  candidate prices for request $r$ by all vehicles that can serve the request. \\

\noindent \textbf{Mean Field Q-Learning, MFQL: }
In problems with a large number of agents with actions that can be aggregated in a meaningful way, mean field Multi-Agent Reinforcement Learning methods (MARL)  methods~\cite{pmlr-v80-yang18d} have been employed successfully. In these methods, interactions between agent populations are approximated as interactions between a single agent and the average influence of neighboring agents. We adapt this broad idea in computing candidate prices for requests in the neighborhood.

The key elements of the learning problem are as follows. Each vehicle is defined as an \textbf{agent}. The \textbf{observation} space of each vehicle is given by:
$$\omega_v = (l_v, l_{N_v}, R_{v})$$
where $l_v$ corresponds to the location (intersection in the road network) of vehicle $v$, $N_v$ corresponds to the vehicles that are close to $v$,  $l_{N_v}$ corresponds to the location of other vehicles in the neighborhood of $v$ and $R_v$ corresponds to the set of requests that can be served by $v$.  The available \textbf{action} set of each agent, $a_v$ is discrete and can take values from set of price factors for that agent, for e.g., $\mathcal{A}_v = \{ 0.8,0.9,1.0,1.1,1.2\}$.

The \textbf{state transition} of an agent is determined by the matching phase (of Section~\ref{matching}) and will take it to the location of the assigned request.  Like with state transition, \textbf{reward} for an agent, $\mathcal{J}_v$ is determined by the outcome of matching and final pricing phase and specifically the request combination, $f$ assigned to vehicle $v$ and the price sensitivity function of the customer. Thus, reward in expectation is $$\sum_{r \in f} \tilde{\mu}^r_v \cdot p^r(\tilde{\mu}^r_v)$$  
However, for a specific request and a customer, it will depend on whether the customer accepted the price. 

The action or price factor ${a}_{v}$ for each vehicle $v$ is a discrete categorical variable represented as the one-hot encoding. State  provides supply demand imbalance in the neighborhood of the current agent. Actions (price factors) of the neighbor agents provide signal of the supply demand imbalance in the areas around the neighborhood. Specifically, we calculate mean action of the neighbor agents, $N_v$ of the vehicle $v$. The average response is ${\overline{a}}_{N_v} = \frac{1}{|N_{v}|}\sum_{k \in N_v}^{}\mspace{2mu} a_{k}$.  Assuming that vehicle $v$ is only affected by the actions of neighboring vehicles, the Q function becomes $Q_{v}\left( \omega_{v},{\overline{a}}_{N_{v}},a_{v} \right)$.

For action selection, the Boltzmann softmax selector is used to obtain the final action probability:  
\begin{equation}
\begin{split}
    a_v & \sim \pi_{v}( \cdot \mid \omega_{v},{\bar{a}}_{N_v}), \\
    \pi_{v}( \! a_{v} \! \mid \! \omega_{v},{\bar{a}}_{N_v}\!) & \!=\! \frac{\exp{\left( \beta \cdot Q_{v}\left( \omega_{v}, a_{v},{\bar{a}}_{N_v} \right) \right)}}{\sum_{a_{v}' \in \mathcal{A}_v} \! \exp \! \left( \beta \cdot Q_{v}\left( \omega_{v},a_{v}',{\bar{a}}_{N_{v}} \right) \right)} 
\end{split}
\end{equation} 
where the $\mathcal{A}_v$ is the action space of vehicle (agent) $v$.

Similar to Q-learning, the mean field Q-function is updated as follows:
\begin{equation}
    Q^{t+1}_{v}\!\left(\!\omega, a_{v}, \bar{a}_{v}\!\right)\!\!=\!(\!1-\alpha\!) Q^{t}_{v}\left(\omega, a_{v}, \bar{a}_{v}\right)+\alpha\!\left[\mathcal{J}_{v}\!\!+\!\!\gamma {V_{\text{MF}}}^{t}_{v}\left(\omega^{\prime}\right)\!\right]
\end{equation}
where 
\begin{equation}
    {V_{\text{MF}}}^{t}_{v}\left(\omega^{\prime}\right) = 
    \sum_{a_{v}} \pi^{t}_{v} \mathbb{E}_{\bar{a}_{v}\left(\boldsymbol{a}_{-v}\right) \sim \pi^{t}_{-v}}\left[Q^{t}_{v}\left(\omega^{\prime}, a_{v}, \bar{a}_{v}\right)\right]
\end{equation}
where the $-v$ is the other agent except agent $v$.
% The update of dynamic pricing strategy is shown in algorithm~\ref{alg:}

\subsection*{Matching RL}
\label{matching}
After computing the candidate prices, we employ future aware (consider matches that will enable better matches in the future) matching to compute the values for different matches given the prices computed in the first step. In this case, we employ an approximate dynamic program as transition uncertainty is external to the action taken. That is to say, uncertainty is associated with customer requests arising at a period and this is not dependent on the action taken. Therefore, we employ an Approximate Dynamic Program (ADP) to exploit this structure, as has been exploited in previous works~\cite{shah2020neural}. 

ADP is similar to a Markov Decision Problem (MDP) with the key difference that the transition uncertainty is extrinsic to the system and not dependent on the action. The ADP problem for the matching problem for each vehicle, $v$ is formulated as follows :
\squishlist
\item[${S}_v$]: The state of a vehicle considers the location and requests that are in the neighborhood.  
\item[${A}_v$]: At each time step there are a large number of requests arriving to the taxi service provider, however for an individual vehicle only a small number of such requests is reachable. The feasible set of request combinations for each vehicle $v$ at time $t$, $\mathcal{F}^v_t$ is given by:
\begin{equation}
\begin{split}
        \mathcal{F}^{t}_{v} = & \left\{ f_{v} \mid f_{v} \in \cup_{c' = 0}^{c^{v}}{\mathcal{\lbrack U}\rbrack^{c'}},\text{PickUpDelay}( f_{v},v) \leq \tau \right.\  \\
        &\left. \ \text{DetourDelay}( f_{v},v) \leq \lambda \right\} \nonumber
\end{split}
\end{equation}
where
$c_{v}$ is the maximum capacity of vehicle, and $c' \leq c_{v}$ is the optional capacity of vehicle $v$. We use $\mathcal{\lbrack U}\rbrack^{c'}$ as a set of permutations and combinations of optional orders under optional capacity ${c'}$.
$\cup_{c^{'} = 0}^{c_{v}}{\mathcal{\lbrack U}\rbrack^{c'}}$ is the union of permutations and combinations under all optional capacities, including the empty set when ${c'}=0$.

$x^{t,f}_v$ is the decision variable that indicates whether vehicle $v$ takes action $f$ (a combination of requests) at a decision epoch $t$. 

%\text{\em At most one request combination, $f$ for each vehicle $i$}
%\text{\em At most one vehicle $i$ assigned to a request $j$}

\item[${\xi}$]: denotes the exogenous information -- the source of randomness in the system. This would correspond to the user requests or demands. $\xi^t$ denotes the exogenous information at time $t$. 
\item[${T}_v$]: denotes the transitions of vehicle state. In an ADP, the system evolution happens as: $$(s_{v}^0,x_v^{0},\tilde{s}_{v}^{0},\xi_{1},s^{1}_v,x^{1}_v,\tilde{s}^{1}_v,\cdots,s^{t}_v,x_v^{t},\tilde{s}_v^{t},\cdots)$$ where $s^{t}_v$ denotes the pre-decision state at decision epoch $t$ for vehicle $v$ and $\tilde{s}_{v}^{t}$ denotes the post-decision state~\cite{powell2007approximate}. The transition from state $s^{t}_v$ to $s^{t+1}_v$ depends on the action vector $x_v^{t}$ and the exogenous information $\xi^{t+1}$. Therefore, 
\begin{align}
    s_v^{t+1} &= {T}_v(s^{t}_v,x^{t}_v,\xi^{t+1}); &
    \tilde{s}^{t}_{v} = \tilde{T}_v(s^{t}_v,x_v^{t}); \nonumber\\
    s_v^{t+1} &= {T}_v^{\xi}(\tilde{s}_v^{t},\xi_{t+1})
\end{align}
It should be noted that $\tilde{T}_v(.,.)$ is deterministic as uncertainty is extrinsic to the system.  
\item[$o_v$]: denotes the revenue for vehicle v considering the candidate prices provided by pricing RL. The $o_v^f(\mu_v^f)$ is vehicle $v$'s revenue which can be obtained by selecting the request combination $f$ given candidate prices, $\mu_v^f$ from pricing RL. The total revenue for vehicle $v$ is
\begin{equation}
    o_v(s_v, f) = o_{v}^f (\mu^{f}_{v})= \sum_{r\in f} \mu^r_v
    \label{equ:objective_fun}
\end{equation}
where $\mu_r^v$ is the candidate price provided by pricing RL for request $r$. 
\squishend

Bellman equations over the joint value of all agents can be provided as follows:
\begin{align}
    V(s^{t}) = \max_{x^{t} \in \mathcal{F}^{t}} (o(s^{t},x^{t}) + \gamma \mathbb{E}[V(s^{t+1})|s^{t},x^{t},\xi^{t+1}])
    \label{eqn:bellman}
\end{align}
where $\gamma$ is the discount factor. Using post-decision state, this expression breaks down nicely:
\begin{align}
    V(s^{t}) &= \max_{x^{t} \in \mathcal{F}^{t}} (o(s^{t},x^{t}) + \gamma {V}(\tilde{s}^{t}));\nonumber\\
    {V}(\tilde{s}^{t}) &= \mathbb{E}[V(s^{t+1})|\tilde{s}^{t},\xi^{t+1}] \label{eqn:post-decision}
\end{align}
For the matching problem, Shah {\em et al.}~\cite{shah2020neural} introduced a two-step decomposition of the joint value function in the second equation above that converts it into a linear combination over individual value functions associated with each vehicle. This allows us to get around the \textit{combinatorial explosion of the post-decision state of all vehicles. } NeurADP thus has the joint value function : 
\begin{equation}V(\tilde{s}^{t}) = \sum_{v} V_{v}([\tilde{s}_{v}^{t}, s^{t}_{\text{-}v}])\label{eqn:decomp}\end{equation}
We then evaluate these individual $V_{v}$ values for all possible $\tilde{s}^{t}_{v}$ from the individual value neural network and pass it to the matching integer linear optimization problem indicated in Figure~\ref{fig:2}.

\subsection*{Centralized Matching: Linear Integer Optimization}

Given the values for different matches for each vehicle from Matching RL, we compute the best joint matching strategy over all available vehicles and customer requests. The key goal of this module is to ensure matching constraints are satisfied, i.e., (a) each vehicle, $v$ can only be assigned at most one request combination, $f$; (b) at most one vehicle, $v$ can be assigned to a request $r$; and (c) a vehicle, $v$ can be either assigned or not assigned to a request combination.\\
 $x_{t}^{v,f}$ is set to 1 if vehicle $v$ takes action $f$ at time $t$. Formally, these three constraints can be specified as follows:
\begin{align}
 & \sum_{f \in {\mathcal F}_{v}^{t}} x_{v}^{t,f} = 1 ::: \forall v ;; \sum_{v} \sum_{f \in {\mathcal F}_{v}^{t};r \in f} x_{v}^{t,f} \leq  1 ::: \forall r \nonumber\\
& x_{v}^{t,f} \in \{0,1\} ::: \forall v,f  \label{cons:a1}
\end{align}

 The myopic objective over all agents is given by:
\begin{equation}
   o(s^t, x^t) = {\sum_{v}\sum_{f \in \mathcal{F}^{t}_{v}}}o^{t,f}_{v} (\mu_v^f) \cdot x^{t}_{v,f}
\end{equation}
This myopic objective decomposes over the individual vehicles.  Similarly, from Equation~\ref{eqn:decomp}, we have the future value decomposing over individual values. Combining these two equations, we have the  following overall Integer Linear Programming (ILP) considering the long term impact: 
\begin{align}
    \max_{\mathbf{x}} & \hspace{0.1in} o(s^{t},x^{t}) + V(\tilde{s}^{t}) \nonumber\\
    \text{s.t.} & \textit{ constraints in (\ref{cons:a1}) are satisfied}
    % \sum_{f \in {\mathcal F}_{v}^{t}} x_{v}^{t,f} = 1 ::: \forall v \nonumber\\
    % &\sum_{v} \sum_{f \in {\mathcal F}_{v}^{t};r \in f} x_{v}^{t,f} \leq  1 ::: \forall r \nonumber\\
    % & x_{v}^{t,f} \in \{0,1\} ::: \forall v,f  \nonumber
\end{align}
Before solving this integer linear optimization, we retrieve values for each vehicle from the individual neural network.

\section*{Experimental Results}

We now describe the experimental setup and the performance results of our overall approach on a benchmark simulation that is based on a city scale real dataset. 

\subsection*{Setup}

\subsubsection{Dataset Description}
We perform our experiments on the demand distribution from the publicly available New York Yellow Taxi Dataset \cite{yellowtaxi}. The experimental setup is similar to the setup used by \cite{shah2020neural}. Street intersections are used as the set of locations ${\mathcal L}$. They are identified by taking the street network of the city from openstreetmap using osmnx with 'drive' network type \cite{boeing2017osmnx}. Nodes that do not have outgoing edges are removed, i.e., we take the largest strongly connected component of the network. The resulting network has 4373 locations (street intersections) and 9540 edges. The travel time on each road segment of the street network is taken as the daily mean travel time estimate computed using the method proposed in \cite{santi2014quantifying}.

Similar to previous work, we only consider the street network of Manhattan as a majority ($\sim$75\%) of requests have both pickup and drop-off locations within it. The dataset contains data about past customer requests for taxis at different times of the day and different days of the week. From this dataset, we take the following fields: (1) Pickup and drop-off locations (latitude and longitude coordinates) - These locations are mapped to the nearest street intersection. (2) Pickup time - This time is converted to appropriate decision epoch based on the value of $\Delta$.  The dataset contains on an average 322714 requests in a day (on weekdays) and 19820 requests during peak hour.

We evaluate the approaches over 24 hours on different days starting at midnight and take the average value over 5 weekdays (4 - 8 April 2016) by running them with a single instance of the initial random location of taxis~\footnote{All experiments are run on 60 core - 3.8GHz Intel Xeon C2 processor and 240GB RAM. The algorithms are implemented in python and optimisation models are solved using CPLEX 20.1}. Our approach is trained using the data for 8 weekdays (23 March - 1 April 2016) and it is validated on 22nd March 2016. For the experimental analysis, we consider that all vehicles have identical capacities. 

\subsubsection{Simulator}

In standard supervised learning problems, the data is stationary for the algorithm. Our 2 layered Reinforcement learning is different from this learning paradigm. It needs continuous interaction with the environment to learn the pricing and matching strategies. Using real-world data to build traffic simulators is a common method in all previous works~\cite{Alonso-Mora462,shah2020neural,lesmana2019balancing,Lowalekar2019ZACAZ}. More details on the simulator provided in Appendix C. 
We use the price sensitive function obtained from Uber data~\cite{Yan2019} to determine the probability of customer acceptance. 
\begin{equation}
    p_{r}(\mu^r_v) = \frac{1}{1 + e^{\frac{0.67 \cdot \mu^r_v}{{\mu^r_0}} - 1.69}}
\label{equ:uber_sen}
\end{equation}
where $\mu^r_0$ is the base price, and  $\mu^r_v$ is the actual price.\\

% \subsubsection{(1) Customers choose whether to take the car or not}\label{passengers-choose-whether-to-take-the-car-or-not}
\subsection*{Baselines}\label{sec:baseline}
To demonstrate the utility of our joint pricing and matching framework, we compared against baselines that optimize pricing and matching individually. Specifically, in these baselines, we replaced the component (pricing/matching) that is not being optimized with existing methods:\\
\noindent (-) To illustrate the role of mean field in our framework, we use Q-learning as a pricing strategy for comparison.\\
\noindent (-) We also set up a baseline pricing algorithm using fixed price which obtains the price through travel distance and time.\\
\noindent (-) To make the algorithm match the vehicles with more reasonable bids, we set up an ADP with the expectation revenue of accepting orders. On the basis of formula \ref{equ:objective_fun} multiplied by an acceptance probability, the expected return can be obtained.  Expected return of the vehicle $v$ is:
\begin{equation}
    o^{t,f}_{v} (\mu^{t,f}_v) = \sum_{r\in f} a^{t}_{v} \cdot \mu_{v}^{t,r} \cdot p_{r}( \mu^{t,r}_v )
    \label{equ:expect_return}
\end{equation}

We combined the components, and the abbreviations of the different approaches are shown in Table~\ref{tab:comparison}.
\begin{table}[t]
\centering
%\resizebox{.95\columnwidth}{!}{
{\small \begin{tabular}{|l l l|}
    \hline
    Name & Pricing & Matching \\
    \hline
    M $\&$ N-E & Mean field  & NeurADP-Exp return-Eqn \ref{equ:expect_return} \\
    M $\&$ N-N & Mean field  & NeurADP-Norm return-Eqn \ref{equ:objective_fun} \\    
    M $\&$ IR & Mean field  & Immediate reward  \\
    Q $\&$ N-E & Q-learning & NeurADP-Exp return-Eqn \ref{equ:expect_return} \\
    Q $\&$ N-N & Q-learning & NeurADP-Norm return-Eqn \ref{equ:objective_fun} \\
    F $\&$ N-E & Fixed price & NeurADP-Exp return \\
    F $\&$ IR & Fixed price & Immediate reward \\
    \hline
\end{tabular}}
\caption{List of all approaches. Fixed price is a myopic pricing approach. Immediate reward is a myopic matching approach. NeurADP is a future aware matching approach. Mean Field is the future aware pricing approach. }
\label{tab:comparison}
\end{table}

\begin{table}[tbp]
\centering
%\resizebox{.95\columnwidth}{!}{
{\small \begin{tabular}{|l|ll|ll|ll|}
    \hline
    \multirow{2}{*}{Name} & \multicolumn{6}{c|}{Result}   \\
    \cline{2-7}
                & \multicolumn{2}{c|}{1000v}   & \multicolumn{2}{c|}{1500v}   & \multicolumn{2}{c|}{2000v} \\
     \cline{2-7}           
             & 2c    & 4c  & 2c  & 4c   & 2c  & 4c\\
    \hline
    M $\&$ N-E    &  9.46	&  5.09	  & 4.74	&  1.75	  &   1.22	&  1.20 \\
    M $\&$ N-N    &  9.35	&  3.06	  & 4.73	&  1.27	  &   0.98	&  1.14 \\
    M $\&$ IR     &6.10	    &-0.32	  &2.75	    & 1.10	  &   1.34	&  0.56  \\
    \hline
    Q $\&$ N-E    &  8.52	&  4.65	  & 3.96	&  1.30	  &   0.71	&  0.66  \\
    Q $\&$ N-N    &  -3.30	&  2.48	  & 1.87	&  -4.94  &   0.36	&  0.02  \\
    F $\&$ N-E    &  0.00   & 0.00    &  0.00   &  0.00   &   0.00  &  0.00   \\
    F $\&$ IR     & -1.20	&-5.11	  &-1.70	&  0.37	  & -0.16   &  -0.01 \\
    \hline
\end{tabular}}
\caption{The impact of different pricing and matching strategies on the results. We show the results of overall revenue for different numbers of vehicles with 1000v referring to 1000 vehicles and  2c referring to a vehicle capacity of  2. 
The values in the table are calculated as the percentage increase relative to the F $\&$ N-E baseline.}

\label{tab:pricing_impact}
\end{table}
\begin{table}[tbp]
\setlength\tabcolsep{2pt}
\centering
%\resizebox{.95\columnwidth}{!}{
{\small \begin{tabular}{l|lllll}
    \hline
    \multirow{2}{*}{} & \multicolumn{5}{c}{Result}   \\
    \cline{2-6}
     & F$\&$N-E    & M$\&$N-E  & M$\&$N-N  & Q$\&$N-E  & Q$\&$N-N \\
    \hline
                Uber        &0.00	&5.09	&3.06	&4.65	&2.48\\
                \hline
                Conscious   &0.00	&17.55	&4.45	&-0.45	&-47.11\\
                \hline
                Very Conscious       &0.00	&15.67	&-34.24	&12.41	&-38.50\\
                % Conscious & & & & & \\
                \hline
\end{tabular}}
\caption{Impact of price sensitivity on the performance. The values in the table are calculated as the percentage increase relative to the F $\&$ N-E baseline. We set the number of vehicles at 1000 and capacity at 4.}
\label{tab:price_sen}
\end{table}
\begin{table}[h]
\centering
%\resizebox{.95\columnwidth}{!}{
{\small \begin{tabular}{lll}
    \hline
    % \multirow{2}{*}{Revenue} & \multicolumn{2}{c}{Number of Vehicles}   \\
    % \cline{2-3}
                & F $\&$ IR   & M $\&$ N-E\\
    \hline
                $\sim$ 200K             &570v 2c	    &530v 2c\\
                $\sim$ 250K             &750v 2c	    &670v 2c\\
                $\sim$ 300K             &940v 2c	    &840v 2c\\
                $\sim$ 350K             &1140v 2c      &990v 2c\\
                $\sim$ 400K             &1380v 2c	    &1180v 2c\\
    \hline
\end{tabular}}
\caption{Number of vehicles required by different algorithms to reach the income level.
}
\label{tab:sust}
\end{table}
\begin{table}[h]
\setlength\tabcolsep{4pt}
\centering
%\resizebox{.95\columnwidth}{!}{
{\small \begin{tabular}{lllc}
    \hline
    % \multirow{2}{*}{} & \multicolumn{2}{c}{Different Algorithm} & \multirow{2}{*}{Save Distance}   \\
    % \cline{2-3}
                                     & F $\&$ IR   & M $\&$ N-E & Save Distance\\
    \hline
                1000v 2c             &11.27	    &11.18  &0.78\%\\
                1000v 4c             &11.27	    &11.09  &1.70\%\\
                1500v 2c             &11.18	    &10.98  &1.85\%\\
                1500v 4c             &11.08     &10.69  &3.50\%\\
                2000v 2c             &10.78	    &10.44  &3.16\%\\
                2000v 4c             &10.56	    &9.38   &11.13\%\\
    \hline
\end{tabular}}
\caption{Average kilometers traveled by all vehicles in one hour (18:00-19:00).
}
\label{tab:savedist}
\end{table}
\subsection*{Results: Efficiency, Sustainability}
\begin{figure}[t]
    \centering
    \includegraphics[width=3.4in,height=1.6in]{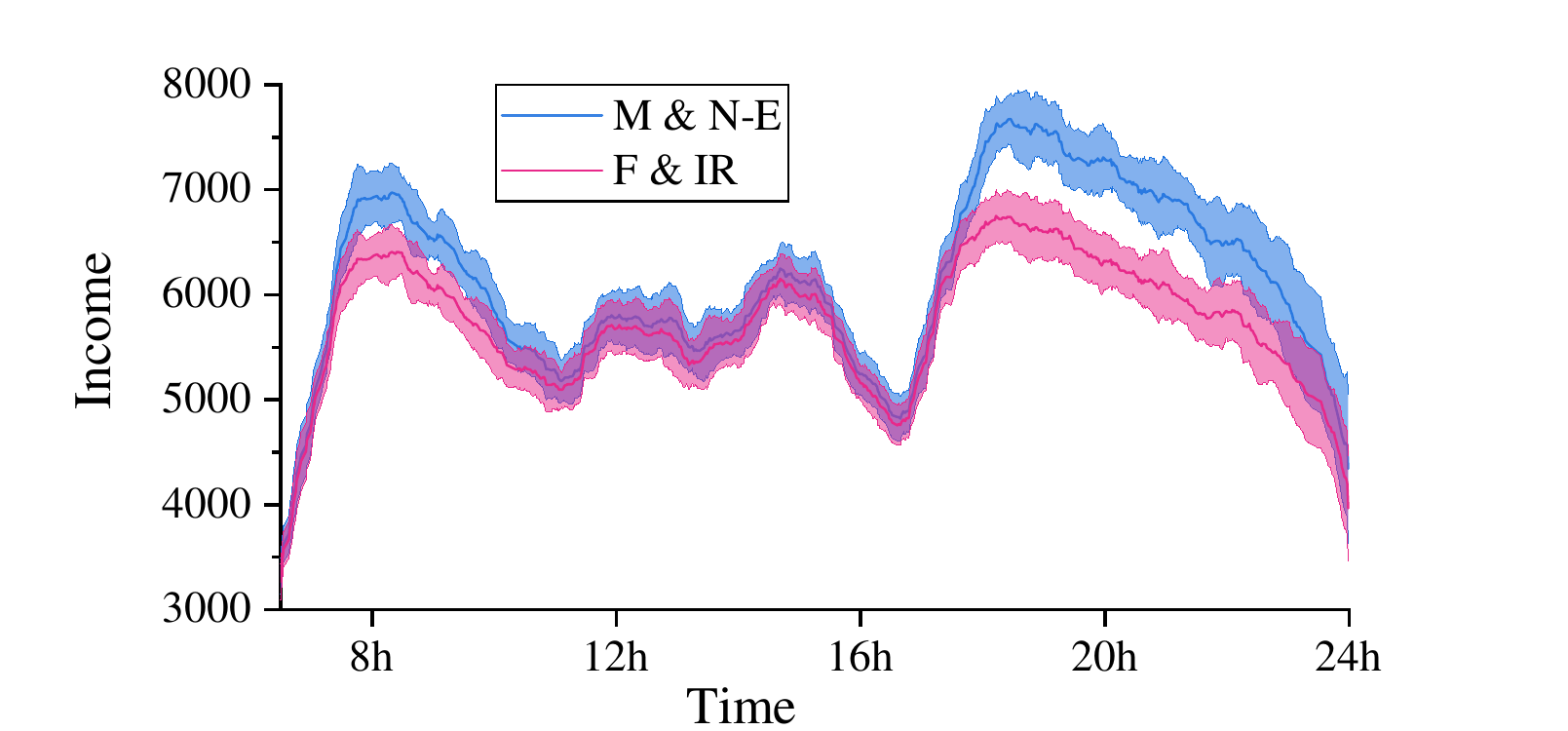}
    \caption{The curve represents the total revenue (mean and one standard deviation over 5 days) of all vehicles at 60-second intervals. We set the number of vehicles at 1000 and capacity at 4.
    }
    \label{fig:income}
\end{figure}
% \begin{figure}[h]
%     \centering
%     \includegraphics[width=0.8\linewidth]{curves_picture/request1000v4c.eps}
%     \caption{Number of requests. The curve represents the number of requests at the interval of 60s. 1000v 4c 300 delay.}
%     \label{fig:request}
% \end{figure}
To illustrate the performance of our proposed framework, comparisons are made along different dimensions. While we tried other peak time periods as well with similar results, we show the results for the evening peak period. Results in all tables are averaged over 5 runs and repeated over the testing days. The first metric we use to compare them is revenue value, which is the total revenue for the selected period. The second metric we use is the number of vehicles required to achieve a fixed revenue level. The decision epoch duration is set as 60 seconds. We utilized F $\&$ N-E as the baseline strategy, as this performed the worst among all approaches where at least one of pricing or matching was future aware. 

\noindent Here are the key observations:\\
\noindent (*) Our best future aware matching and pricing strategy, i.e., M $\&$ N-E provides up to 9\% improvement over a method that employs future aware matching but myopic pricing, i.e., F $\&$ N-E. There is improvement across all settings and can be attributed to the simultaneous matching and pricing optimization in our approach. As expected, this improvement reduces as more higher capacity vehicles are available.  \\
\noindent (*) Our best future aware matching and pricing strategy, i.e., M $\&$ N-E provides up to 6\% improvement over a method that employs future aware pricing but myopic matching, i.e., M $\&$ IR. Improvement is consistent and reduces as higher capacity vehicles are available.\\
\noindent (*)  Given a fixed matching strategy, our mean field pricing strategy due to considering neighbor strategies provides consistent improvement (up to ~12\%) over a Q-learning based pricing strategy.\\
\noindent (*) Using a myopic pricing and matching strategy performs the worst among all approaches. \\
\noindent (*) Using a sample based estimate for matching rewards, i.e., N-N strategy provided slightly lower performance than using expected rewards, i.e., N-E strategy, especially when we used mean field pricing strategy. 

\noindent \textbf{Impact on environment:} To demonstrate that our approach is more sustainable, we fixed the overall revenue number and computed the number of vehicles  required to achieve that revenue for different approaches. Table~\ref{tab:sust} shows the results. We observe that our M $\&$ N-E approach provides a reduction of up to 14.5\% (and on average 11.2\%)in the number of vehicles over a myopic pricing and matching approach. 

We also computed the average number of kilometers traveled by all vehicles during the peak period. Table~\ref{tab:savedist} provides these results. We were able to provide a reduction of up to 11.13\% and on average a reduction of 3.8\%. 

\noindent \textbf{Impact of price sensitivity, $p$}:
Customers in different cities can have different levels of price sensitivity. The uber one we have shown so far was for one city. When we made the sensitivity function more price conscious (which is something that can potentially happen in developing or under-developed countries), the impact of our approach was very significant as shown in Table~\ref{tab:price_sen}. We achieve this price consciousness by scaling the coefficient of the final price and the constant term in Equation~\ref{equ:uber_sen} by a factor of 10 which makes the transition from 'accept' to 'not accept' more dramatic. Once again, M $\&$ N-E approach clearly outperformed other approaches and the improvement was up to 17\% over the baseline. 

\section*{Conclusion}

The emergence of Uber, LYFT, DiDi and other applications has improved service for customers while having higher utilization for taxis. Most existing work has focused on separately optimizing for the matching and pricing problems, and this can reduce the effectiveness, specifically in on-demand ride pooling systems. We have provided a novel 2 layered Reinforcement Learning approach with a centralized optimization for simultaneously optimizing pricing and matching. Our approach consistently improves over existing baselines and in the best case achieves an impressive 17\% improvement in revenue and 14\% reduction in number of vehicles for a city scale taxi dataset. This is a big improvement considering that even a 1\% improvement in revenue is considered a big improvement in similar transportation problems by the industry~\cite{Xu2018}. 

\section*{Acknowledgments}
% Acknowledgments.\\
% This research is supported in part by the China Scholarship Council (No. 202006060229).
This research/project is supported by the National Research Foundation Singapore and DSO National Laboratories under the AI Singapore Programme (AISG Award No: AISG2-RP-2020-017) and China Scholarship Council (No: 202006060229).

\bibliography{sample-base}

\begin{thebibliography}{25}
\providecommand{\natexlab}[1]{#1}

\bibitem[{Alonso-Mora et~al.(2017)Alonso-Mora, Samaranayake, Wallar, Frazzoli,
  and Rus}]{Alonso-Mora462}
Alonso-Mora, J.; Samaranayake, S.; Wallar, A.; Frazzoli, E.; and Rus, D. 2017.
\newblock On-demand high-capacity ride-sharing via dynamic trip-vehicle
  assignment.
\newblock \emph{Proceedings of the National Academy of Sciences}, 114(3):
  462--467.

\bibitem[{Banerjee, Johari, and Riquelme(2015)}]{3_p}
Banerjee, S.; Johari, R.; and Riquelme, C. 2015.
\newblock Pricing in Ride-Sharing Platforms: A Queueing-Theoretic Approach.
\newblock In \emph{Proceedings of the Sixteenth ACM Conference on Economics and
  Computation}, EC '15, 639. New York, NY, USA: Association for Computing
  Machinery.
\newblock ISBN 9781450334105.

\bibitem[{Banerjee, Johari, and Riquelme(2016)}]{banerjee2016dynamic}
Banerjee, S.; Johari, R.; and Riquelme, C. 2016.
\newblock Dynamic pricing in ridesharing platforms.
\newblock \emph{ACM SIGecom Exchanges}, 15(1): 65--70.

\bibitem[{Banerjee, Riquelme, and Johari(2015)}]{banerjee2015pricing}
Banerjee, S.; Riquelme, C.; and Johari, R. 2015.
\newblock Pricing in ride-share platforms: A queueing-theoretic approach.
\newblock \emph{Available at SSRN 2568258}.

\bibitem[{Bimpikis, Candogan, and Saban(2019)}]{bimpikis2019spatial}
Bimpikis, K.; Candogan, O.; and Saban, D. 2019.
\newblock Spatial pricing in ride-sharing networks.
\newblock \emph{Operations Research}, 67(3): 744--769.

\bibitem[{Boeing(2017)}]{boeing2017osmnx}
Boeing, G. 2017.
\newblock OSMnx: New methods for acquiring, constructing, analyzing, and
  visualizing complex street networks.
\newblock \emph{Computers, Environment and Urban Systems}, 65: 126--139.

\bibitem[{Chen et~al.(2019{\natexlab{a}})Chen, Jiao, Qin, Tang, Li, An, Zhu,
  and Ye}]{chen2019inbede}
Chen, H.; Jiao, Y.; Qin, Z.; Tang, X.; Li, H.; An, B.; Zhu, H.; and Ye, J.
  2019{\natexlab{a}}.
\newblock InBEDE: Integrating Contextual Bandit with TD Learning for Joint
  Pricing and Dispatch of Ride-Hailing Platforms.
\newblock In \emph{2019 IEEE International Conference on Data Mining (ICDM)},
  61--70. IEEE.

\bibitem[{Chen et~al.(2019{\natexlab{b}})Chen, Shen, Tang, and
  Zuo}]{chen2019dispatching}
Chen, M.; Shen, W.; Tang, P.; and Zuo, S. 2019{\natexlab{b}}.
\newblock Dispatching through pricing: modeling ride-sharing and designing
  dynamic prices.
\newblock In \emph{Proceedings of the 28th International Joint Conference on
  Artificial Intelligence}, 165--171. AAAI Press.

\bibitem[{Lesmana, Zhang, and Bei(2019)}]{lesmana2019balancing}
Lesmana, N.~S.; Zhang, X.; and Bei, X. 2019.
\newblock Balancing efficiency and fairness in on-demand ridesourcing.
\newblock In \emph{Advances in Neural Information Processing Systems},
  5309--5319.

\bibitem[{Lowalekar, Varakantham, and Jaillet(2018)}]{14_p}
Lowalekar, M.; Varakantham, P.; and Jaillet, P. 2018.
\newblock Online spatio-temporal matching in stochastic and dynamic domains.
\newblock \emph{Artificial Intelligence}, 261: 71--112.

\bibitem[{Lowalekar, Varakantham, and Jaillet(2019)}]{Lowalekar2019ZACAZ}
Lowalekar, M.; Varakantham, P.; and Jaillet, P. 2019.
\newblock ZAC: A Zone Path Construction Approach for Effective Real-Time
  Ridesharing.
\newblock In \emph{ICAPS}.

\bibitem[{Ma, Fang, and Parkes(2019)}]{ma2019spatio}
Ma, H.; Fang, F.; and Parkes, D.~C. 2019.
\newblock Spatio-temporal pricing for ridesharing platforms.
\newblock In \emph{Proceedings of the 2019 ACM Conference on Economics and
  Computation}, 583--583.

\bibitem[{Ma, Zheng, and Wolfson(2013)}]{ma2013t}
Ma, S.; Zheng, Y.; and Wolfson, O. 2013.
\newblock T-share: A large-scale dynamic taxi ridesharing service.
\newblock In \emph{Data Engineering (ICDE), 2013 IEEE 29th International
  Conference on}, 410--421. IEEE.

\bibitem[{NYYellowTaxi(2016)}]{yellowtaxi}
NYYellowTaxi. 2016.
\newblock New York Yellow Taxi DataSet.
\newblock \url{http://www.nyc.gov/html/tlc/html/about/trip_record_data.shtml}.
\newblock Accessed: 2021-12-04.

\bibitem[{{\"O}zkan(2020)}]{ozkan2020joint}
{\"O}zkan, E. 2020.
\newblock Joint pricing and matching in ride-sharing systems.
\newblock \emph{European Journal of Operational Research}.

\bibitem[{Powell(2007)}]{powell2007approximate}
Powell, W.~B. 2007.
\newblock \emph{Approximate Dynamic Programming: Solving the curses of
  dimensionality}, volume 703.
\newblock John Wiley \& Sons.

\bibitem[{Santi et~al.(2014)Santi, Resta, Szell, Sobolevsky, Strogatz, and
  Ratti}]{santi2014quantifying}
Santi, P.; Resta, G.; Szell, M.; Sobolevsky, S.; Strogatz, S.~H.; and Ratti, C.
  2014.
\newblock Quantifying the benefits of vehicle pooling with shareability
  networks.
\newblock \emph{Proceedings of the National Academy of Sciences}, 111(37):
  13290--13294.

\bibitem[{Shah, Lowalekar, and Varakantham(2020)}]{shah2020neural}
Shah, S.; Lowalekar, M.; and Varakantham, P. 2020.
\newblock Neural Approximate Dynamic Programming for On-Demand Ride-Pooling.
\newblock In \emph{The Thirty-Fourth {AAAI} Conference on Artificial
  Intelligence}, 507--515. {AAAI} Press.

\bibitem[{Shah, Lowalekar, and Varakantham(2022)}]{shah2022icaps}
Shah, S.; Lowalekar, M.; and Varakantham, P. 2022.
\newblock Joint Pricing and Matching for City-Scale Ride-Pooling.
\newblock \emph{Proceedings of the International Conference on Automated
  Planning and Scheduling}, 32(1): 499--507.

\bibitem[{Uber(2018)}]{uberblog}
Uber. 2018.
\newblock Uber Matching Solution.
\newblock \url{https://marketplace.uber.com/matching}.
\newblock Accessed: 2021-12-04.

\bibitem[{Xu et~al.(2018)Xu, Li, Guan, Zhang, Li, Nan, Liu, Bian, and
  Ye}]{Xu2018}
Xu, Z.; Li, Z.; Guan, Q.; Zhang, D.; Li, Q.; Nan, J.; Liu, C.; Bian, W.; and
  Ye, J. 2018.
\newblock Large-Scale Order Dispatch in On-Demand Ride-Hailing Platforms: A
  Learning and Planning Approach.
\newblock In \emph{Proceedings of the 24th ACM SIGKDD International Conference
  on Knowledge Discovery and Data Mining}, KDD '18, 905–913.

\bibitem[{Yan et~al.(2019)Yan, Zhu, Korolko, and Woodard}]{Yan2019}
Yan, C.; Zhu, H.; Korolko, N.; and Woodard, D. 2019.
\newblock Dynamic pricing and matching in ride-hailing platforms.
\newblock \emph{Naval Research Logistics ({NRL})}, 67(8): 705--724.

\bibitem[{Yang et~al.(2018)Yang, Luo, Li, Zhou, Zhang, and
  Wang}]{pmlr-v80-yang18d}
Yang, Y.; Luo, R.; Li, M.; Zhou, M.; Zhang, W.; and Wang, J. 2018.
\newblock Mean Field Multi-Agent Reinforcement Learning.
\newblock In \emph{Proceedings of the 35th International Conference on Machine
  Learning (ICML)}, volume~80, 5571--5580.

\bibitem[{Zheng, Chen, and Ye(2018)}]{zheng2018order}
Zheng, L.; Chen, L.; and Ye, J. 2018.
\newblock Order dispatch in price-aware ridesharing.
\newblock \emph{Proceedings of the VLDB Endowment}, 11(8): 853--865.

\bibitem[{Özkan(2020)}]{jointpandm}
Özkan, E. 2020.
\newblock Joint pricing and matching in ride-sharing systems.
\newblock \emph{European Journal of Operational Research}, 287(3): 1149--1160.

\end{thebibliography}

\end{document}